\documentclass[12pt]{article}
\usepackage[a4paper,margin=2.35cm]{geometry}
\usepackage[T1]{fontenc}
\usepackage[utf8]{inputenc}
\usepackage[english]{babel}
\usepackage{amsmath,amsthm,mathtools}
\usepackage{newtxtext,newtxmath}
\usepackage{graphicx}
\usepackage{enumitem}
\usepackage{microtype}
\usepackage[hidelinks]{hyperref}

\title{Tensor Data Scattering and the Impossibility\\
of Slicing Theorem}
\author{Wuming Pan\\
Sichuan University, Chengdu 610065, China\\
\textit{Correspondence:} \texttt{panwuming@scu.edu.cn}}
\date{}

\makeatletter
\renewcommand{\maketitle}{%
  \begin{flushleft}
    {\fontsize{16}{19}\selectfont\bfseries\@title\par}
    \vspace{1.0em}
    {\normalsize\bfseries\@author\par}
  \end{flushleft}
  \vspace{0.8em}}
\makeatother

\begin{document}
\maketitle

\begin{abstract}
This paper proposes a standard way to represent sparse tensors. A broad theoretical framework for tensor data scattering methods used in various deep learning frameworks is established. This paper presents a theorem that is very important for performance analysis and accelerator optimization for implementing data scattering. The theorem shows how the impossibility of slicing happens in tensor data scattering. A sparsity measuring formula is provided, which can effectively indicate the storage efficiency of sparse tensor and the possibility of parallelly using it. A Python reference implementation is provided as ancillary material with this arXiv submission.
\end{abstract}


\section{Introduction}

\noindent Most data used in AI and big data analysis is multidimensional in nature. Storing them in multi-dimension way, known as tensors, is more efficient than matrixes or two-dimensional arrays, hence tensor is gaining more and more importance in AI computing. Some current computing architectures support parallel computing along two or three dimensions on data, such as CUDA architectures [1], this facilitates computing related to tensors. However, many matrixes and tensors used in AI computing contain fewer data than their capacities, and they are often stored sparsely. For example, sparse attentions, which attract recent research interests, always result in sparse matrixes [2]. Sparsely stored matrixes or tensors are difficult to use the hardware features of machine learning accelerators. 

\noindent Sparse tensors are typically stored with an array of indices and an array of values at corresponding indices, which is like what sparse matrixes are. Sparse tensors can be more easily found inherent duplication on their storing structures than sparse matrixes, hence they can be stored and used in different ways. Some APIs in deep learning frameworks, including TensorFlow and pyTorch, are developed to store and use sparse tensors in such ways, they are often called scatter APIs [3,4]. If a sparse tensor has duplicated storing structures along some dimensions, it can be transported and used in computing parallelly. However, there hasn't common way to scattering tensor data. We will analyze the reasons for those difficulties in this paper. Though there are many tensor operations having been provided, such as NumPy array operations [5] and tubal-rank tensor operations [6], we still define new operations on tensors in this paper. 

\noindent Next in this paper we will define tensors and related notions mathematically. In section 3, we define a useful operation called picks. Then, in the section 4, we define the tensor variator and its provision tensor that are critical for tensor data scattering. In the fifth section we discuss the uncertainty brought about by the applying of the variator, which is the source of uncertainty in the scattering of tensor data. In section 6, we define scattering and suggest how scatterings can be sliceable. In section 7, we define the x-sparse representation of sparse tensors and the x-scattering operation, suggest how to count sparsity of a sparse tensor, and show how the TensorFlow and pyTorch style APIs are mocked. The last section is the conclusion.

\section{Tensor}

\noindent To strictly discuss these problems, we need argue them mathematically. Other than in programming language, the syntax representations are not able to make two mathematical objects different. All mathematical objects eventually should be embodied in the set theory. We need start from elementary mathematic objects, such as the set  $\mathbb{N}$ of natural numbers, the real number set $\mathbb{R}$, products of sets, mappings, and functions between sets, etc. We use ${\mathbb{N}}_m$ to denote the set of $m$ nonnegative integers from 0 to $m-1$, i.e.   ${\mathbb{N}}_m=\left\{0,1,\cdots ,m-1\right\}$. Specifically, we think that ${\mathbb{N}}_0=\ \left\{\ \right\}=\emptyset $. We also use ${\mathbb{N}}_{h:k}$ to denote the set $\left\{h,h+1,\cdots ,k-1\right\}$ where $h\le k$ and both are integers, possibly negative. 

\noindent A tuple is an element in a Cartesian product of some sets. For a tuple $t$, $\ell \left(t\right)$ denotes its length, i.e., the number of sets comprise product set which $t$ is in. A tuple with length 1 is a number. A tuple with length 0 is not a number, just represented as $\left(\ \right)$. We treat tuples as they can be concatenated with operator + as python tuples. For example:$\left(1,2,3\right)+4=\left(1,2,3,4\right)$

\noindent In deep learning area, a tensor is a multidimensional data array. Tensor's data elements can be accessed through their indices. An index of a data element is a tuple. 

\noindent \textbf{Definition 1.} \textit{A }\textbf{\textit{tensor }}\textit{is a function}

\[
E:\prod^{k-1}_{i=0}{{\mathbb{N}}_{m_i}}\to \mathbb{R}
\]
\noindent \textit{The tuple }$S=\left(m_0,m_1,\cdots ,m_{k-1}\right)$\textit{ is called the }\textbf{\textit{shape}}\textit{ of tensor }$E$\textit{, denoted with }${\mathrm{s}}_E$\textit{. And the elements in }$\prod^{k-1}_{i=0}{{\mathbb{N}}_{m_i}}$\textit{ is called }\textbf{\textit{indices}}\textit{ of }$E$\textit{. The notation }${\mathbb{I}}_S$\textit{ stands for the set of all indices of shape }$S$\textit{, i.e.,}$\ \ {\mathbb{I}}_S=\prod^{k-1}_{i=0}{{\mathbb{N}}_{m_i}}$\textit{ , and the notation }${\mathbb{I}}_E$\textit{ stands for the set of all indices of tensor }$E$\textit{ as well. For any }$\left(j_0,j_1,\cdots ,j_{k-1}\right)\in \ {\mathbb{I}}_S$\textit{, we simply write }$E\left(\left(j_0,j_1,\cdots ,j_{k-1}\right)\right)$\textit{ as }$E\left[j_0,j_1,\cdots ,j_{k-1}\right]$\textit{. The number of total elements in }$E$\textit{ is called its }\textbf{\textit{size}}\textit{, denoted as }${\mathrm{\Pi }}_E$\textit{ or }${\mathrm{\Pi }}_S$\textit{. The operator }$\mathrm{t}$\textit{ change a one-dimension tensor to a tuple}. \textit{A tensor with shape }$\left(\ \right)$\textit{ is an empty tensor denoted as }$\left[\ \right]$\textit{. We define that }$\mathrm{t}\left(\left[\ \right]\right)=\left(\ \right)$\textit{.}

\noindent To represent a tensor, we use $\left[\ \right.$ and $\left.\ \right]$ to bracket tensor data. Unlike representing a matrix, the subtensors are arranged horizontally or vertically in the same dimensions, but not at both directions.  For example, a tensor $E0$ has the shape $\left(3,3,2\right)$, and $E0$ is represented as

\[
\resizebox{\textwidth}{!}{$
E\_0=\left[ \begin{array}{ccc}
\left[\left[ \begin{array}{cc}
0 & 0 \end{array}
\right]\left[ \begin{array}{cc}
0 & 1 \end{array}
\right]\left[ \begin{array}{cc}
0 & 2 \end{array}
\right]\right] & \left[\left[ \begin{array}{cc}
1 & 0 \end{array}
\right]\left[ \begin{array}{cc}
1 & 1 \end{array}
\right]\left[ \begin{array}{cc}
1 & 2 \end{array}
\right]\right] & \left[\left[ \begin{array}{cc}
2 & 0 \end{array}
\right]\left[ \begin{array}{cc}
2 & 1 \end{array}
\right]\left[ \begin{array}{cc}
2 & 2 \end{array}
\right]\right] \end{array}
\ \right]
$}
\]
\noindent or
\[
E\_0=\left[ \begin{array}{c}
\left[\left[ \begin{array}{cc}
0 & 0 \end{array}
\right]\left[ \begin{array}{cc}
0 & 1 \end{array}
\right]\left[ \begin{array}{cc}
0 & 2 \end{array}
\right]\right] \\ 
\left[\left[ \begin{array}{cc}
1 & 0 \end{array}
\right]\left[ \begin{array}{cc}
1 & 1 \end{array}
\right]\left[ \begin{array}{cc}
1 & 2 \end{array}
\right]\right] \\ 
\left[\left[ \begin{array}{cc}
2 & 0 \end{array}
\right]\left[ \begin{array}{cc}
2 & 1 \end{array}
\right]\left[ \begin{array}{cc}
2 & 2 \end{array}
\right]\right] \end{array}
\right]
\]

\noindent These are different than matrix data arrange format.  

\section{Pick and Slice}

\noindent \textbf{Definition 2}. \textit{A }\textbf{\textit{pick}}\textit{ is a finite integer function }$p:{\mathbb{N}}_n\to \mathbb{N}$\textit{ for some nonnegative integer }$n$\textit{.We also use }$\ell \left(p\right)$\textit{ to denote }$n$\textit{.} \textit{The pick }$p$\textit{ is }\textbf{\textit{smooth}}\textit{ if only if }$p$\textit{ monotonically maps consecutive numbers to consecutive numbers. We define that }$\mathrm{c}\left(p\right)=p\left({\mathbb{N}}_n\right)$\textit{. We use }$max\left(p\right)$\textit{ to denote the maximum element in }$\mathrm{c}\left(p\right)$\textit{. If }$p\left({\mathbb{N}}_n\right)={\mathbb{N}}_n$\textit{, then we call }$p$\textit{ is a }\textbf{\textit{shuffle}}\textit{. If }$p$\textit{ is one to one, then we call }$p$\textit{ is }\textbf{\textit{simple}}\textit{. Let }$I$\textit{ be a tuple with }$\ell \left(I\right)\ge max\left(p\right)$\textit{ , }$p\left(I\right)$\textit{ is a tuple }$J$\textit{ such that }$J\left[i\right]=I[p\left(i\right)]$\textit{. If }$p$\textit{ is simple, then it has an inverse partial function }$p^{-1}:\mathrm{c}\left(p\right)\to {\mathbb{N}}_n$\textit{. We define picks }${\mathrm{i}}_n:{\mathbb{N}}_n\to {\mathbb{N}}_n$\textit{ as  }${\mathrm{i}}_n\left(j\right)=j$\textit{ and call them}\textbf{\textit{ identity picks}}\textit{ with rank }$n$\textit{. We define picks }${\mathrm{i}}_{n:m}:{\mathbb{N}}_{n:m}\to {\mathbb{N}}_{n:m}$\textit{ as  }${\mathrm{i}}_{n:m}\left(j-n\right)=j$\textit{ and call them}\textbf{\textit{ identity picks}}\textit{ with rank }$n$\textit{ to }$m$\textit{.}

\noindent Sometimes a pick is used as a projection into indices along a dimension, at other times a pick can be used to select dimensions of a tensor. A pick can be written as a one-dimensional integer tensor. 

\noindent \textbf{Definition 3}. \textit{Given a simple pick }$p$, \textit{let }${{\mathbb{I}}_S}/{p}$\textit{ be the set of some subsets of }${\mathbb{I}}_S$\textit{ such that for any }$C\in {{\mathbb{I}}_S}/{p}$\textit{, there is a }$I_C\in {\mathbb{I}}_S$\textit{, and}

\noindent $C=\left\{K\left|p\left(\ I_C\right)=\right.p\left(K\right),K\in {\mathbb{I}}_S\right\}$\textit{and vice versa. Let }$E$\textit{ be a tensor with shape }$S$\textit{, we call }$E\left(C\right)$\textit{ is a }\textbf{\textit{slice}}\textit{ of }$E$\textit{ }\textbf{\textit{picked}}\textit{ by }$p$. \textit{If }$p=\ {\mathrm{i}}_m$\textit{ for some m, then we call }$E\left(C\right)$\textit{ a }\textbf{\textit{subtensor}}\textit{ of }$E$\textit{, also denoted as }$E\left(C\right)=E\left[p\left(\ I_C\right)\right]$\textit{.}

\noindent For example, let

\[
\begin{aligned}
E\_00&=\left[\left[ \begin{array}{cc}
0 & 0 \end{array}
\right]\left[ \begin{array}{cc}
0 & 1 \end{array}
\right]\left[ \begin{array}{cc}
0 & 2 \end{array}
\right]\right]\\
E\_01&=\left[\left[ \begin{array}{cc}
1 & 0 \end{array}
\right]\left[ \begin{array}{cc}
1 & 1 \end{array}
\right]\left[ \begin{array}{cc}
1 & 2 \end{array}
\right]\right]\\
E\_02&=\left[\left[ \begin{array}{cc}
2 & 0 \end{array}
\right]\left[ \begin{array}{cc}
2 & 1 \end{array}
\right]\left[ \begin{array}{cc}
2 & 2 \end{array}
\right]\right]
\end{aligned}
\]
\noindent Then $E00$, $E01$ and $E02$ are subtensors of a tensor $E0$, which can be represented as
\[
E\_0=\left[ \begin{array}{ccc}
E\_00 & E\_01 & E\_02 \end{array}
\right]=\left[ \begin{array}{ccc}
E\_0\left[0\right] & E\_0\left[1\right] & E\_0\left[2\right] \end{array}
\right]
\]

\section{Tensor Variator and Its Provision Tensor}

\noindent \textbf{Definition 4}. \textit{A }\textbf{\textit{tensor variator}}\textit{ is a map }$T:{\mathbb{I}}_{S_0}\to {\mathbb{I}}_{S_1}$\textit{, it can be used as an operation on tensors. Let }$A$\textit{ be a tensor with shape }$S_0$\textit{, then }$T\left(A\right)$\textit{ is a set of tensors. For any }$B\in T\left(A\right)$\textit{ and }$I\in {\mathbb{I}}_{S_1}$\textit{, there exists a }$J\in T^{-1}\left(T\left(I\right)\right)$\textit{, such that }$B\left[T\left(J\right)\right]=A[J]$.

\noindent A variator can be defined by a tensor.

\noindent \textbf{Definition 5}. \textit{Given a variator }$T:{\mathbb{I}}_{S_0}\to {\mathbb{I}}_{S_1}$\textit{, let }$\mathrm{e}\left(T\right)$\textit{ be a tensor }$E$\textit{ with shape }$S_0+\ell \left(S_1\right)$\textit{ and it is defined by}

\noindent $T\left(I\right)=\ \mathrm{t}\left(E\left[I\right]\right)$\textit{then we call }$T$\textit{ is }\textbf{\textit{provisioned}}\textit{ by }$E$\textit{, and }$E$\textit{ is a }\textbf{\textit{provisioner}}\textit{ of }$T$\textit{.} 

\noindent \textit{Notes and Comments}. $E\left[I\right]$ is a one-dimensional tensor of $E$.

\noindent For example, let $S_1=\left(4,2\right)$ and $S_2=\left(2,2,2,2\right)$, given tensor of shape $\left(4,2,4\right)$

\noindent $E_{\mathrm{1}}\mathrm{=}\left[\  \begin{array}{c}
\left[\  \begin{array}{c}
\left[ \begin{array}{cccc}
0 & 0 & 0 & 0 \end{array}
\right] \\ 
\left[ \begin{array}{cccc}
0 & 0 & 0 & \mathrm{1} \end{array}
\right] \end{array}
\ \right] \\ 
\left[\  \begin{array}{c}
\left[ \begin{array}{cccc}
0 & 0 & \mathrm{1} & 0 \end{array}
\right] \\ 
\left[ \begin{array}{cccc}
0 & 0 & \mathrm{1} & \mathrm{1} \end{array}
\right] \end{array}
\ \right] \\ 
\left[\  \begin{array}{c}
\left[ \begin{array}{cccc}
0 & \mathrm{1} & 0 & 0 \end{array}
\right] \\ 
\left[ \begin{array}{cccc}
0 & \mathrm{1} & 0 & \mathrm{1} \end{array}
\right] \end{array}
\ \right] \\ 
\left[\  \begin{array}{c}
\left[ \begin{array}{cccc}
0 & \mathrm{1} & \mathrm{1} & 0 \end{array}
\right] \\ 
\left[ \begin{array}{cccc}
0 & \mathrm{1} & \mathrm{1} & \mathrm{1} \end{array}
\right] \end{array}
\ \right] \end{array}
\ \right]$a variator $T$ can be defined as
\[T\left(I\right)\mathrm{=}\left(E\mathrm{\_1}\left[I\left(0\right),I\left(\mathrm{1}\right),0\right]\right.,E\mathrm{\_1}\left[I\left(0\right),I\left(\mathrm{1}\right)\mathrm{,1}\right],E\mathrm{\_1}\left[I\left(0\right),I\left(\mathrm{1}\right)\mathrm{,2}\right],\] 
\[\left.E\mathrm{\_1}\left[I\left(0\right),I\left(\mathrm{1}\right)\mathrm{,3}\right]\right)\] 
If $I=\left(3,0\right)$ then $T\left(I\right)=\left(0,1,1,0\right)$.

\section{Nondeterministic of Applying Variator}

\noindent Given a variator $T:{\mathbb{I}}_{S_0}\to {\mathbb{I}}_{S_1}$, let $A$ be a tensor with shape $S_0$, and a tensor $B\in T\left(A\right)$. There are non-deterministic cases when applying a variator on a tensor:  

\begin{enumerate}
\item  For any $B\in T\left(A\right)$, for any $I\in {\mathbb{I}}_{S_0}$, and for any $J\in T^{-1}\left(T\left(I\right)\right)$, that $B\left(T\left(I\right)\right)=A\left(I\right)$ or $B\left(T\left(I\right)\right)=A\left(J\right)$ holds is both possible. If  $A\left(I\right)\neq A\left(J\right)$, then only one is possibly true.

\item  Moreover, for some $K\in {\mathbb{I}}_{{\mathrm{s}}_B}$, possibly $K\notin T\left({\mathbb{I}}_{S_0}\right)$ is true.

\item  The shape of tensor $B$ is not unique. Any shape $S$ with $T\left({\mathbb{I}}_{S_0}\right)\subset {\mathbb{I}}_S$ can be the shape of $B$. Only length of the shape B is definite.
\end{enumerate}

\noindent Now we can investigate the scattering algorithms in deep learning frameworks. Those algorithms can eliminate the second and the third indeterminate problem above.

\section{Scattering}

\subsection{Scatter APIs in Two Popular Deep Learning Frameworks}

\noindent In TensorFlow, the typical scattering API looks like [5]:

tensor\_scatter\_nd\_update(ts, indices, updates, name=None)

\noindent where \textit{ts}, \textit{indices}, \textit{updates} in argument list are all tensors. Use the notions in this paper, the \textit{indices} argument in this API is used to form a provisioner of a variator $T_{indices}:{\mathbb{I}}_{S_0}\to {\mathbb{I}}_{Si}$ where 
\[S_0={\mathrm{i}}_{\ \ell \left({\mathrm{s}}_{indices}\right)-1}\left({\mathrm{s}}_{indices}\right)\] 
and$\ $there are tuples $Si$ and $Su$ such that
\[{\mathrm{s}}_{ts}=S_2=Si+Su\] 
Let $S_1=S_0+Su$

\noindent and $T_1:{\mathbb{I}}_{S_1}\to {\mathbb{I}}_{S_2}$ be a variator such that for any $I\in {\mathbb{I}}_{S_0}$ and  $J\in {\mathbb{I}}_{Su}$
\[T_1\left(I+J\right)=T_{indices}\left(I\right)+J\] 
This API creates a tensor $B$ which is a result of variator $T_1$ being applied on tensor \textit{updates}, such that there is an 
\[I'\in T^{-1}_{indices}\left(T_{indices}\left(I\right)\right)\] 
for each $I\in {\mathbb{I}}_{S_0}$ and
\[ts\left[T_{indices}\left(I\right)\right]=updates\left[I'\right]\] 
And for any

\noindent $K\notin T_1\left({\mathbb{I}}_{S_1}\right)$the identity $B\left(K\right)=ts\left(K\right)$ must holds. 

\noindent In pyTorch, the typical scattering API looks like [6]:

scatter(self, dim, index, src, reduce=None) 

\noindent where \textit{self}, \textit{index}, \textit{src} in argument list are tensors having same shape, while \textit{dim} is an integer. Still use the notions in this paper, the \textit{index} argument in this API is a tensor to form a provisioner $E_{index}$ of a variator $T_{E_{index}}:{\mathbb{I}}_{S_{index}}\to {\mathrm{i}}_1$ where $E_{index}\left[I\right]=\left[index\left[I\right]\right]$. Then we can define a variator $T_{scatter}:{\mathbb{I}}_{S_{index}}\to {\mathbb{I}}_{S_{index}}$ such that

\noindent $T_{scatter}\left(I\right)={\mathrm{i}}_{dim}\left(I\right)+T_{E_{index}}\left(I\right)+{\mathrm{i}}_{dim+1:\ell \left(I\right)}\left(I\right)$And then this API creates a tensor $C$ which is a result of variator $T_{scatter}$ being applied on the tensor $self$, for any $I\in {\mathbb{I}}_{S_1}$, there is an 
\[I'\in T^{-1}_{scatter}\left(T_{scatter}\left(I\right)\right)\] 
such that
\[C\left(T_{scatter}\left(I\right)\right)=src\left(I'\right)\] 
For any $K\notin {\mathbb{I}}_{S_{index}}$, the identity $C\left(K\right)=self\left(K\right)$ must holds. 

\subsection{Defining Scattering}

\noindent A scattering is a tensor variator being applied on a tensor $A$ and the result tensor $B$ is restricted by a tensor $X$.   

\noindent \textbf{Definition 6.} \textit{A }\textbf{\textit{scattering}}\textit{ is a triple }$e=\left(T,A,X\right)$\textit{, where }$T:{\mathbb{I}}_{S_0}\to {\mathbb{I}}_{S_1}$\textit{ is a tensor variator, }$A$\textit{ and }$X$\textit{ are two tensors. A result of scattering }$e$\textit{ is a result }$B$\textit{ of }$T$\textit{ being applied on }$A$\textit{, and for any }$I$\textit{, if }$I\in T\left({\mathbb{I}}_{S_0}\right)$\textit{ then there is some }$J\in T^{-1}\left(I\right)$\textit{ such that }$B\left[I\right]=A\left[J\right]$\textit{ holds; if }$I\notin T\left({\mathbb{I}}_{S_1}\right)$\textit{, then }$B\left[I\right]=X[I]$\textit{ holds.}

\noindent Since a variator can be represented by a tensor, a scattering is also defined by a triple $\left(E,A,X\right)$ of tensors $E$, $A$ and $X$. For an instance, using tensor $E\_1$ in section 4 to provision a variator, given

\[
A\_1=\left[ \begin{array}{cccc}
\left[ \begin{array}{cc}
1 & 2 \end{array}
\right] & \left[ \begin{array}{cc}
3 & 4 \end{array}
\right] & \left[ \begin{array}{cc}
5 & 6 \end{array}
\right] & \left[ \begin{array}{cc}
7 & 8 \end{array}
\right] \end{array}
\right]
\]
\noindent and
\[
X\_1=\left[ \begin{array}{c}
\left[ \begin{array}{cc}
\left[ \begin{array}{c}
\left[ \begin{array}{cc}
0 & 0 \end{array}
\right] \\ 
\left[ \begin{array}{cc}
0 & 0 \end{array}
\right] \end{array}
\right] & \left[ \begin{array}{c}
\left[ \begin{array}{cc}
0 & 0 \end{array}
\right] \\ 
\left[ \begin{array}{cc}
0 & 0 \end{array}
\right] \end{array}
\right] \end{array}
\right] \\ 
\left[ \begin{array}{cc}
\left[ \begin{array}{c}
\left[ \begin{array}{cc}
0 & 0 \end{array}
\right] \\ 
\left[ \begin{array}{cc}
0 & 0 \end{array}
\right] \end{array}
\right] & \left[ \begin{array}{c}
\left[ \begin{array}{cc}
0 & 0 \end{array}
\right] \\ 
\left[ \begin{array}{cc}
0 & 0 \end{array}
\right] \end{array}
\right] \end{array}
\right] \end{array}
\right]
\]
\noindent Then the result of applying $E\_1\ $ on $A\_1$ into $X\_1$ is a tensor
\[B\_1=\left[ \begin{array}{cc}
\left[ \begin{array}{c}
\left[ \begin{array}{c}
\left[ \begin{array}{cc}
1 & 2 \end{array}
\right] \\ 
\left[ \begin{array}{cc}
3 & 4 \end{array}
\right] \end{array}
\right] \\ 
\left[ \begin{array}{c}
\left[ \begin{array}{cc}
5 & 6 \end{array}
\right] \\ 
\left[ \begin{array}{cc}
7 & 8 \end{array}
\right] \end{array}
\right] \end{array}
\right] & \left[ \begin{array}{c}
\left[ \begin{array}{c}
\left[ \begin{array}{cc}
0 & 0 \end{array}
\right] \\ 
\left[ \begin{array}{cc}
0 & 0 \end{array}
\right] \end{array}
\right] \\ 
\left[ \begin{array}{c}
\left[ \begin{array}{cc}
0 & 0 \end{array}
\right] \\ 
\left[ \begin{array}{cc}
0 & 0 \end{array}
\right] \end{array}
\right] \end{array}
\right] \end{array}
\right]\] 
The result of scattering is also indeterministic. 

\subsection{Sliceable Scattering}

\noindent In some case, when we scatter data from a source tensor into a target tensor, we can replace a slice of target tensor with a slice of source tensor, and the two slices have the same shape. If both slices can be specified with last few dimensions of shapes of both tensors, we can call such scatter operation sliceable. We extend this idea here:

\noindent \textbf{Definition 7}. \textit{A tensor variator }$T:{\mathbb{I}}_{S_0}\to {\mathbb{I}}_{S_1}$\textit{ is called }\textbf{\textit{sliceable}}\textit{, if and only if there are two picks }$p_1$\textit{ and }$p_2$\textit{,  and the three conditions are satisfied:}

\begin{enumerate}
\item  $p_1$\textit{ is simple}.

\item  \textit{For any }$I\in {\mathbb{I}}_{S_0}$\textit{,}$\ \mathrm{c}\left(p_1\right)\cap \mathrm{c}\left(p_2\right)=\emptyset $.

\item  \textit{There is a variator }$T'$\textit{ such that for any }$I\in {\mathbb{I}}_{S_0}$\textit{,  }$T\left(I\right)=T'\left(p_1\left(I\right)\right)+p_2\left(I\right)$\textit{ holds}.
\end{enumerate}

\noindent $p_2$\textit{ is called a }\textbf{\textit{sliceable end pick}}\textit{ of }$T$\textit{. If there does not exist a nonempty sliceable pick for }$T$\textit{, we call it }\textbf{\textit{not sliceable}}\textit{.}

\noindent \textbf{Theorem 1}. \textbf{(Slice Theorem).} \textit{Conditions being as in above definition, for any}$C\in {{\mathbb{I}}_{S_0}}/{p_1}$\textit{ there is}
\[T\left(C\right)\in {{\mathbb{I}}_{S_1}}/{{\mathrm{i}}_{\ \ell \left(S_1\right)-\ell \left(p_2\right)}}\] 
\textit{Proof.} Clear.

\noindent \textit{Notes and Comments}. When $\mathrm{c}\left(p_1\right)\cap \mathrm{c}\left(p_2\right)\neq \emptyset $, the conclusion of this theorem no longer holds. In such case, there would be some $C'\in {{\mathbb{I}}_{S_1}}/{{\mathrm{i}}_{\ \ell \left(S_1\right)-\ell \left(p_2\right)}}$ and $T\left(C\right)\subseteq C'$.

\noindent Let's see an example. The tensor variator with a provision tensor

\[
E\mathrm{\_2=}\left[\  \begin{array}{cc}
\left[\  \begin{array}{c}
\left[\  \begin{array}{c}
\left[ \begin{array}{cccc}
0 & 0 & 0 & 0 \end{array}
\right] \\ 
\left[ \begin{array}{cccc}
0 & 0 & 0 & \mathrm{1} \end{array}
\right] \end{array}
\ \right] \\ 
\left[\  \begin{array}{c}
\left[ \begin{array}{cccc}
0 & 0 & \mathrm{1} & 0 \end{array}
\right] \\ 
\left[ \begin{array}{cccc}
0 & 0 & \mathrm{1} & \mathrm{1} \end{array}
\right] \end{array}
\ \right] \end{array}
\ \right] & \left[\  \begin{array}{c}
\left[\  \begin{array}{c}
\left[ \begin{array}{cccc}
\mathrm{1} & \mathrm{1} & 0 & 0 \end{array}
\right] \\ 
\left[ \begin{array}{cccc}
\mathrm{1} & \mathrm{1} & 0 & \mathrm{1} \end{array}
\right] \end{array}
\ \right] \\ 
\left[\  \begin{array}{c}
\left[ \begin{array}{cccc}
\mathrm{1} & \mathrm{1} & \mathrm{1} & 0 \end{array}
\right] \\ 
\left[ \begin{array}{cccc}
\mathrm{1} & \mathrm{1} & \mathrm{1} & \mathrm{1} \end{array}
\right] \end{array}
\ \right] \end{array}
\ \right] \end{array}
\mathrm{\ }\right]
\]
\noindent This variator is sliceable because there are two picks
\[
p\mathrm{\_2\_1=}\left[0\right], \qquad
p\mathrm{\_2\_2=}\left[ \begin{array}{cc}
\mathrm{1} & \mathrm{2} \end{array}
\right],
\]
and a variator $T\mathrm{'\_2}$ whose provision tensor is
\[\mathrm{e}\left(T\mathrm{'\_2}\right)\mathrm{=}\left[\mathrm{\ } \begin{array}{c}
\left[ \begin{array}{cc}
0 & 0 \end{array}
\right] \\ 
\left[ \begin{array}{cc}
\mathrm{1} & \mathrm{1} \end{array}
\right] \end{array}
\mathrm{\ }\right]\] 

\section{Sparse Tensor with X-Sparse Representation}

\subsection{The Limitations in Current Scattering APIs}

\noindent Not any kinds of scattering are implemented in deep learning framework currently. There are limitations in current scattering APIs:

\begin{enumerate}
\item  The scattering APIs are not compatible with each other. For example, TensorFlow scattering APIs is sliceable, whereas pyTorch scattering APIs is almost not sliceable. Both cannot efficiently mock the behavior of each other. 

\item  A tensor variator properly represented as weakly sliceable can be stored efficiently. However, weakly sliceable scattering is not implemented in any deep learning frameworks. We will define weakly sliceable later.
\end{enumerate}

\noindent Now we design a scattering algorithm which can directly address weak sliceable possibility of scattering. A weakly sliceable scattering API can incorporate functionalities of both TensorFlow and pyTorch scattering APIs. 

\subsection{X-Sparse Tensor}

\noindent A x-variator $T$ is decorated with three picks and a variator.

\noindent \textbf{Definition 8}. \textit{A tensor variator }$T:{\mathbb{I}}_{S_0}\to {\mathbb{I}}_{S_1}$\textit{ is called a }\textbf{\textit{x-variator}}\textit{, if and only if there are three picks }$p_1$\textit{, }$p_2$\textit{, }$p$\textit{ and a variator }$f:{\mathbb{I}}_{p_1\left(S_0\right)}\to {\mathbb{I}}_{S_2}$\textit{, and for any }$I\in {\mathbb{I}}_{S_0}$\textit{ there is }
\[T\left(I\right)=p\left(f\left(p_1\left(I\right)\right)+p_2\left(I\right)\right)\] 
\textit{We call the tuple }$\left(f,p_1,p_2,p\right)$\textit{ a}\textbf{\textit{ x-representation}}\textit{, or simply a }\textbf{\textit{representation}}\textit{, of  }$T$\textit{. If }$max\left(p\right)>\ell \left(S_2\right)$\textit{, then the tuple is called a }\textbf{\textit{normal}}\textit{ representation. If }$\mathrm{c}\left(p_1\right)\cap \mathrm{c}\left(p_2\right)\neq \emptyset $\textit{, and there is some }$i\in \mathrm{c}\left(p_1\right)\cap \mathrm{c}\left(p_2\right)$\textit{ and some }$j\in p^{-1}_2\left(i\right)$\textit{}
\[j+\ell \left(S_2\right)\in \ \mathrm{c}\left(p\right)\] 
\textit{We call  }$\left(f,p_1,p_2,p\right)$\textit{ is an}\textbf{\textit{ entangled}}\textit{ representation of }$T$.

\noindent Any variator $T:{\mathbb{I}}_{S_0}\to {\mathbb{I}}_{S_1}$ evidently have a x-variator representation 

\noindent $\left(T,\ {\mathrm{i}}_{\ell \left(S_0\right)},{\mathrm{i}}_0,{\mathrm{i}}_{\ell \left(S_1\right)}\right)$That is, for any\textit{ }$I\in {\mathbb{I}}_{S_0}$ we have
\[T\left(I\right)={\mathrm{i}}_{\ell \left(S_1\right)}\left(T\left({\mathrm{i}}_{\ell \left(S_0\right)}\left(I\right)\right)+{\mathrm{i}}_0\left(I\right)\right)\] 
\textbf{Corollary 1}. \textit{Given any variator }$T$\textit{, it is a x-variator.}

\noindent The representation of a variator as a x-variator is not unique.

\noindent \textbf{Theorem 2}. \textbf{(Impossibility of Slicing Theorem).} \textit{Given a variator }$T$\textit{, if for any representation }$\left(f,p_1,p_2,p\right)$\textit{ of }$T$\textit{, that the condition }$\mathrm{c}\left(p_1\right)\cap \mathrm{c}\left(p_2\right)\neq \emptyset $\textit{ holds implies that the representation is entangled, then} $T$\textit{ is not sliceable.}

\noindent \textit{Proof.} It is clear.

\noindent \textbf{Definition 9}. \textit{We say a tensor variator }$T:{\mathbb{I}}_{S_1}\to {\mathbb{I}}_{S_2}$\textit{ is }\textbf{\textit{weakly}}\textit{ }\textbf{\textit{sliceable}}\textit{, if and only if  }$T$\textit{ has a normal representation  }$\left(f,p_1,p_2,p\right)$\textit{, such that  }$\mathrm{c}\left(p_1\right)\cap \mathrm{c}\left(p_2\right)=\emptyset $\textit{.}

\noindent Let's see another example. The variator

\noindent $E\mathrm{\_3=}\left[\  \begin{array}{cc}
\left[\  \begin{array}{c}
\left[\  \begin{array}{c}
\left[ \begin{array}{ccc}
0 & 0 & 0 \end{array}
\right] \\ 
\left[ \begin{array}{ccc}
\mathrm{1} & 0 & \mathrm{1} \end{array}
\right] \end{array}
\ \right] \\ 
\left[\  \begin{array}{c}
\left[ \begin{array}{ccc}
0 & \mathrm{1} & 0 \end{array}
\right] \\ 
\left[ \begin{array}{ccc}
\mathrm{1} & \mathrm{1} & \mathrm{1} \end{array}
\right] \end{array}
\ \right] \end{array}
\ \right] & \left[\  \begin{array}{c}
\left[\  \begin{array}{c}
\left[ \begin{array}{ccc}
0 & 0 & 0 \end{array}
\right] \\ 
\left[ \begin{array}{ccc}
\mathrm{1} & 0 & \mathrm{1} \end{array}
\right] \end{array}
\ \right] \\ 
\left[\  \begin{array}{c}
\left[ \begin{array}{ccc}
0 & \mathrm{1} & 0 \end{array}
\right] \\ 
\left[ \begin{array}{ccc}
\mathrm{1} & \mathrm{1} & \mathrm{1} \end{array}
\right] \end{array}
\ \right] \end{array}
\ \right] \end{array}
\mathrm{\ }\right]$is weakly sliceable. Because there are picks
\[p\mathrm{\_3\_1=}\left[0\right]p\mathrm{\_3\_2=}\left[\mathrm{1}\right]\] 
$p\mathrm{\_3=}\left[ \begin{array}{ccc}
0 & \mathrm{2} & \mathrm{1} \end{array}
\right]$and a variator $f\mathrm{\_3}$ whose provision tensor is 
\[\mathrm{e}\left(f\mathrm{\_3}\right)\mathrm{=}\left[\  \begin{array}{c}
\left[ \begin{array}{cc}
0 & 0 \end{array}
\right] \\ 
\left[ \begin{array}{cc}
\mathrm{1} & \mathrm{1} \end{array}
\right] \end{array}
\ \right]\] 
\textbf{Definition 10}. \textit{Given a tensor variator }$T:{\mathbb{I}}_{S_0}\to {\mathbb{I}}_{S_1}$\textit{ which has a normal representation  }$\left(f,p_1,p_2,p\right)$\textit{, where }$p$\textit{ is a shaffle. }Let $F$ be the provisioner tensor of the \textit{variator }$f:{\mathbb{I}}_{p_1\left(S_0\right)}\to {\mathbb{I}}_{S_2}$, \textit{and let }$V$\textit{ be a tensor with shape }$S_0$\textit{, then  }$\left(F,p_1,p_2,p,V\right)$\textit{ is called a }\textbf{\textit{x-sparse tensor representation}}\textit{, or simply }\textbf{\textit{x-sparse tensor}}\textit{.  A}\textbf{\textit{ x-scattering}}\textit{ is a binary  }$e=\left(A,X\right)$\textit{, where }$A=\left(F,p_1,p_2,p,V\right)$\textit{ is the} \textit{x-sparse tensor, }$X$\textit{ is a tensor. A result of x-scattering }$e$\textit{ is a tensor }$B$\textit{ defined as for any }$J$\textit{, if }$J\in T\left({\mathbb{I}}_{S_0}\right)$\textit{, then there is some }$I\in {\mathbb{I}}_{S_0}$\textit{, such that}
\[J=p\left(J_0\right)\] 
\textit{where}
\[J_0=p_1\left(I\right)+p_2\left(I\right)\] 
\textit{ and  }$B\left[J\right]=V\left(I\right)$\textit{; if }$J\in {\mathbb{I}}_{S_B}$\textit{ and }$J\notin T\left({\mathbb{I}}_{S_0}\right)$\textit{, then }$B\left[J\right]=X[J]$\textit{ holds}

\noindent The result tensor of a x-scattering also cannot be certainly determined.

\noindent We provide a Python reference implementation of x-scattering as ancillary material with this arXiv submission and call it the scatterX API.

\subsection{Counting Sparsity and Analyzing Performance}

\noindent A dense tensor $E$ with shape $S$ has an x-sparse tensor representation
\[
\left(\ \left[\ \right],{\mathrm{i}}_0,{\mathrm{i}}_n,{\mathrm{i}}_n,E\right).
\]
Once we randomly remove a few elements from $E$ and get a sparse tensor $E'$, then $E'$ has an x-sparse representation
\[
\left(\ indices,{\mathrm{i}}_1,{\mathrm{i}}_0,{\mathrm{i}}_n,V\right),
\]
where $indices$ is a provisioner of a variator $T:{\mathbb{I}}_{{\mathrm{i}}_1}\to {\mathbb{I}}_{{\mathrm{i}}_n}$ and $V$ is a one-dimensional tensor that contains elements of $E'$. Thus, the inner variator in an x-sparse tensor indicates the efficiency of storing sparse indices.

\noindent \textbf{Definition 11}: \textit{Given a x-sparse tensor  }$X=\left(F,p_1,p_2,p,V\right)$\textit{, the }\textbf{\textit{sparsity}}\textit{ of the x-sparse tensor is defined as}
\[{\mathrm{a}}_X=\frac{{\mathrm{\Pi }}_F}{{\mathrm{\Pi }}_V*\ell \left(p\right)}\] 
Now we can count the sparsity of former examples:
\[\ {\mathrm{a}}_{\left(\ \left[\ \right],{\mathrm{i}}_0,{\mathrm{i}}_n,{\mathrm{i}}_n,E\right)}=0\] 
\[{\mathrm{a}}_{\left(\ indices,{\mathrm{i}}_1,{\mathrm{i}}_0,{\mathrm{i}}_n,V\right)}\approx 1\] 
The sparsity is 1 means that the x-sparse tensor hardly can be parallelly used. The x-sparse tensor has smaller sparsity will have high possibility to be parallelly used.

\subsection{Mocking Current Scattering APIs}

\noindent The counterpart scattering of the TensorFlow scatter API as in section 6 has a x-scattering representation
\[\left(\left(indices,p_1,p_2,p,updates\right),ts\right)\] 
where
\[p={\mathrm{i}}_{\ \ell \left({\mathrm{s}}_{ts}\right)}\] 
\[n=\ell \left({\mathrm{s}}_{indices}\right)-1\] 
\[p_1={{\mathrm{i}}_n}_{\ }\] 
\[m=\ell \left({\mathrm{s}}_{updates}\right)\] 
\[p_2={{\mathrm{i}}_{n:m}}_{\ }\] 
The sparsity
\[{\mathrm{a}}_{\left(indices,p_1,p_2,p,updates\right)}=\frac{{\mathrm{\Pi }}_{indices}}{{\mathrm{\Pi }}_{updates}*\ell \left(p\right)}\le \frac{1}{{\mathrm{\Pi }}_S*\ell \left(p\right)}\le \frac{1}{\ell \left(p\right)}\] 
where
\[S={\mathrm{i}}_{{\mathrm{s}}_{indices}:m}\left({\mathrm{s}}_{updates}\right)\] 
It can be any number smaller than 1. Whereas the counterpart scattering of the pyTorch scatter API as in section 6 has a x-scattering representation
\[\left(\left(E_{index},q_1,q_2,q,src\right),self\right)\] 
where
\[q={\mathrm{i}}_{1:\left(dim+1\right)}+0+{\mathrm{i}}_{\left(dim+1\right)\ :\ell \left({\mathrm{s}}_{src}\right)}\] 
\[q_1=q_2={\mathrm{i}}_{\ \ell \left({\mathrm{s}}_{src}\right)}\] 
The sparsity
\[{\mathrm{a}}_{\left(E_{index},q_1,q_2,q,src\right)}=\frac{{\mathrm{\Pi }}_{E_{index}}}{{\mathrm{\Pi }}_{src}*\ell \left(q\right)}=\frac{1}{\ell \left(q\right)}\] 
This means that pyTorch scatter API is not sliceable.

\noindent The key difference of these two kinds of APIs is how the variator in scattering is formed.

\section{Conclusion}

\noindent Tensor data scattering is a kind of task that is difficult to use the hardware features of machine learning accelerators. This article theoretically analyses the reasons for this difficulty. And a general theory and algorithm of tensor data scattering is established in this article. Based on the theories and algorithms in this article, we will be able to implement algorithms that can make better use of accelerator features. Moreover, a standard approach is proposed to represent sparse tensor, which can facilitate parallel computing and data transporting in AI accelerators, and which can also provide a way to efficiently store sparse indices of sparse tensors. A sparsity measuring formula is provided at last section, which can effectively indicate the storage efficiency of sparse tensor and the possibility of parallelly using it. More experiments and comparisons with APIs in other deep learning frameworks remain for future work.

\section*{References}

\begin{sloppypar}
\begin{enumerate}
\item \textbf{ }Soyata, T.: GPU parallel program development using CUDA. CRC Press, Florida (2018).

\item  Child, R., Gray, S., Radford, A., Sutskever, I.: Generating Long Sequences with Sparse Transformers. arXiv:1904.10509 (2019).

\item  TensorFlow API: tf.tensor\_scatter\_nd\_update, \url{https://www.tensorflow.org/api_docs/python/tf/tensor_scatter_nd_update}.

\item  PyTorch Docs: torch.Tensor.scatter\_, \url{https://docs.pytorch.org/docs/stable/generated/torch.Tensor.scatter_.html}.

\item  Harris, C.R., Millman, K.J., van der Walt, S.J.~\textit{et al.}:~Array programming with NumPy.~Nature~585,\textbf{~}357--362 (2020).

\item  Zhang, T., Liu, X., Wang, X., Walid, A.: cuTensor-Tubal: Efficient Primitives for Tubal-Rank Tensor Learning Operations on GPUs, IEEE Transactions on Parallel and Distributed Systems, 31(3), 595--610 (2020).
\end{enumerate}
\end{sloppypar}

\noindent 

\noindent

\end{document}